\documentclass[conference]{IEEEtran}
\IEEEoverridecommandlockouts
\usepackage{cite}
\usepackage{amsmath,amssymb,amsfonts}
\usepackage{graphicx}
\usepackage{textcomp}
\usepackage{tabularx}
\usepackage{xcolor}
\usepackage{textalpha}
\usepackage{multirow}
\usepackage[hyphens]{url}
\usepackage{hyperref}
\hypersetup{colorlinks=true,breaklinks=true}
\usepackage{caption}
\usepackage{subcaption}
\usepackage{comment}
\usepackage{makecell}
\usepackage{pdfpages}

\def\BibTeX{{\rm B\kern-.05em{\sc i\kern-.025em b}\kern-.08em
    T\kern-.1667em\lower.7ex\hbox{E}\kern-.125emX}}
\begin{document}

\title{Understanding the Effect of Smartphone Cameras on Estimating Munsell Soil Colors from Imagery\thanks{\copyright IEEE, accepted to publish in DICTA 2022 proceedings}}

\author{\IEEEauthorblockN{Ricky Sinclair}
\IEEEauthorblockA{\textit{School of Computing, Mathematics and Engineering} \\
\textit{Charles Sturt University}\\
NSW, Australia \\
rickysinclair@gmail.com}
\and
\IEEEauthorblockN{ Muhammad Ashad Kabir\textsuperscript{1,2}}
\IEEEauthorblockA{\textsuperscript{1}\textit{School of Computing, Mathematics and Engineering} \\
\textsuperscript{2}\textit{Gulbali Institute for Agriculture, Water and Environment}\\
\textit{Charles Sturt University}\\
NSW, Australia \\
akabir@csu.edu.au}
}

\maketitle

\begin{abstract}
The Munsell soil color chart (MSCC) is an important reference for many professionals in the area of soil color analysis. Currently, the functionality to identify Munsell soil colors (MSCs) automatically from an image is only feasible in laboratories under controlled conditions. 
To support an app-based solution, this paper explores three research areas including: (i) identifying the most effective color space, (ii) establishing the color difference calculation method with the highest accuracy and (iii) evaluating the effects of smartphone cameras on estimating the MSCs. 
The existing methods that we have analysed have returned promising results and will help inform other researchers to better understand and develop informed solutions. This study provides both researchers and developers with an insight into the best methods for automatically predicting MSCs. Future research is needed to improve the reliability of results under differing environmental conditions.
\end{abstract}


\section{Introduction}\label{}
Many theories have developed over the years in attempts to better explain how colors work and how best to calculate and describe color differences. In 1913, Albert Munsell introduced the Atlas of the Munsell color system, arranging color into the tristimulus of hue, value, and chroma \cite{munsell1913}. The Munsell color system has enabled professionals to bridge the disciplines of art and science and is the basis of many professional applications today such as food science \cite{giese2000}, dentistry \cite{Chang2012}, printing \cite{allen1921}, painting \cite{cochrane2014}, and soil science \cite{rossel2006}. Soil colors are most easily measured by comparison with a Munsell soil color chart (MSCC), an adaptation of the Munsell color space that includes only the values needed for soil colors, about one-fifth of the complete series of colors. The organisation of the MSCC is the same as the Munsell soil color system hue, value, and chroma (HVC). 

Color is one of the most noticeable features of soil and is very important in the assessment and classification of soil properties. It can also provide valuable insights into the soil environment, often being used for predictively determining soil properties and the land's appropriateness for uses. The prediction of soil color is also of national importance in Australia, as current land management techniques are reducing the levels of nutrients in the soil at alarming rates \cite{liefting_2020}.

Several researchers have developed varying techniques to predict Munsell soil colors (MSCs) from a range of technologies, and the opinions of each vary when it comes to the selection of the best technologies to utilise in this endeavour \cite{fan2017, milotta2020, heil2020, Aitkenhead2020, Kirillova2021}. Whilst current research discusses the method of color detection techniques, the justification in selecting each individual element in these methods is lacking bringing into question the processes for automatic MSCs predictions.

In this paper, our primary focus is understanding and analysing the various color spaces (i.e., RGB, CIELCh, CIELab, XYZ, and CMYK) and the color difference calculation methods (i.e., CIEDE1976, CIEDE1994, CIEDE2000, and CMC) for the MSCs. Consequently, the color difference calculation methods have examined to enable a working theory for estimating MSCs from smartphone captured images. The effect of different smartphone cameras is also analysed and discussed. The findings from this paper established CIELab is the most appropriate color space using the CIEDE2000 color difference calculation method combined with the D65 illuminant.

\section{Methodology}\label{}

A review of the color spaces used for the conversion of the MSCC and the associated color difference calculation methods will be investigated and evaluated for their effectiveness. This review will enable the researchers to select the best color space and conversion algorithms. 

Furthermore, the effects of smartphone cameras on estimating the MSCs will be explored. More specifically, an analysis of the differences between smartphone cameras and the accuracy of colors and images captured across devices will be explored. Color correction techniques will also be assessed for their effectiveness in increasing the reliability and accuracy of Munsell color identification. This study will focus on being able to accurately and reliably being able to match colors to the MSCC from captured images.

Although, research in the area of Munsell soil color prediction is extensive, there seems to be little justification for the use of the methodologies applied. For example, many researchers focus on the devices \cite{fan2017, heil2020, Kirillova2021} being used rather than the conversion and calculation methods, taking these processes for granted in their results. Fewer studies have compared these methods for their reliability and effectiveness \cite{Aitkenhead2020, Kirillova2021}. There remains the need for an analysis to be undertaken that develops a cohesive solution incorporating a thorough discernment of the devices, color space, color difference calculation methods, effects of smartphone cameras (SPCs), and color correction techniques used in the prediction of Munsell soil color values. These key issues are detailed in the following subsections.

\section{Color spaces for MSCC}\label{}

Mobile phones can only record image colors in RGB. Munsell colors, therefore, need to be interpolated from RGB to Munsell. However, a range of possible color spaces could be utilised as a proxy between RGB, and the Munsell Soil Color Chart, to improve the accuracy of samples. If a medium is needed, then the question remains, which color space is best to act as a medium between RGB and Munsell? The RGB, CMYK, CIEXYZ, CIELCh and CIELab color spaces were selected as possible mediums as this colorimetry data is available from the Nix Pro 2~\footnote{\url{https://www.nixsensor.com/nix-pro/}} which will allow a thorough analysis of the effects that each color space has during the conversion process.

RGB is composed of a tristimulus, in which the hue values move from red through other colors and back to red \cite{Aitkenhead2016}. This makes it difficult to calculate distances from one color to another in the RGB color space. To establish a numerical description of these colors, the categories must be transformed to a value. Each color system has various advantages when converting from a Cartesian color coordinate system to Munsell color descriptors. The choice of the right hue also becomes more difficult the lower the Chroma of a color \cite{Gerharz1988}. Since the distance between the adjacent Hue chart gets ever smaller the nearer they approach the axis. Soils frequently contain subdued colors (i.e. low Chromas), increasing the complexity in the determination of Hue. 

One of the main issues is that these values can be very close to one another, and any slight change in one of the tristimulus results in the hue of the Munsell value changing. This can create problems for the repeatability of results and, therefore, affect the accuracy of the readings. A 3D spatial model of the interpolations between the color space and the MSCC was developed to analyse the most effective color space. What we are looking for here is a clear separation of the `clumping' of results. This `clumping' shows that the different hues in the MSCC are too closely aligned, meaning that any small changes in the tristimulus of a sample, converted to a particular color space, results in a large movement in the MSCC to another hue rather than an adjacent chip in the MSCC.

The MSCC contains the color chips from 14 hues in total. When analysing the potential use of a color space as a medium creating a 3D spatial model for all of these chips would become quite populated and difficult to read. Therefore, the most relevant hues were determined as applied to Australian soils. The Australian national site collation researched the most prominent hue colors for samples collected around Australia \cite{Global2014}. Over 680,000 observations were recorded from topsoils recorded at a depth shallower than 5cm from the surface. This data displays a trend that the primary soil colors in Australia are: 5Y, 2.5Y, 10YR, 7.5YR, 5YR, 2.5YR and 10R. This essentially removes gleys, bright red colors, and whites. This is a strong justification that the research in this area focuses on these hues from the MSCC. 


Munsell notations are not always unambiguous. Apart from the human error and the individual color perception, the determinations allow for some uncertainty due to the closeness of the values, especially when more difficult materials, like soils, are valued. To investigate the relevance of this uncertainty, each color space was explored and discussed as a potential medium for the interpolation process.

The Munsell color space is based on a three-dimensional model in which each color is comprised of the tristimulus of hue (color type), value (lightness/darkness) and chroma (color saturation). Hue, value and chroma are also annotated as (HVC).

The Munsell color space gives us an intuitive designation to express our perception of color and its changes, similar to that of the human eye \cite{Pendleton1951}. However, the subjective nature of Munsell color charts and the limited number of color chips can impose restrictions on precise color measurements \cite{melville1985, xu2019}. Whilst the MSCC remains the standard color space for analysing soil. As discussed earlier, this color space is not natively readable by electronic devices and relies upon varied interpolation methods. In this section, we aim to determine the best color space to store soil samples from a SPC and then compare them to the MSCC color chips via interpolation. 

RGB is the basic color space for digital cameras and computer displays that use red, green, and blue to create the required color. Therefore, the RGB color space is the obvious choice for software systems due to its ease of integration. For this reason, \cite{Aitkenhead2016} argued in favour of the RGB system to be used to convert Munsell soil colors. In the RGB system, all three highly correlated bands determine illumination intensity jointly, which is the major disadvantage of the system \cite{rossel2008}. This means that the conversion of Munsell codes to RGB is not appropriate at certain high values and chroma levels as the range of RGB values is in effect not large enough to cover all of the possible Munsell values. Because of this disadvantage, other color spaces are considered a better choice for Munsell interpolation \cite{heil2020}. 

The RGB color space would be the most convenient color space to use for Munsell soil color comparison as it would require no medium and can be converted directly to match the MSCC. Indeed, \cite{Han2016} recommended using RGB values when utilising a smartphone camera to process color images of soil samples as RGB is the native color space for these devices. Similarly, RGB was utilised by the majority of researchers \cite{Han2016, Kirillova2018, Milotta2018a, Pegalajar2020, schmidt2021, Stiglitz2017, Stiglitz2016b}. However, several studies incorporated a combination of color spaces to determine which method returned the best results \cite{fan2017, Fu2020, Gorthi2021}. \cite{Pegalajar2020} argues that RGB measurements from the mobile phone obtained good results, but using a more stable color space, such as CIELab, might improve the HVC approximations. \cite{fan2017} disagreed with the use of the RGB color space as the standard RGB color space created by a digital camera is device dependent, and the color capturing and processing procedures are influenced by the device. \cite{Milotta2018a} used the Munsell conversion tables by \cite{Centore2012} in which sampled RGB values were been converted to the Munsell color space. These tables rely on fixed conversion pairs RGB to Munsell, and viceversa.

\begin{figure*}[t!]
    \centering
    \begin{subfigure}[b]{.46\textwidth}
    \includegraphics[width=\textwidth]{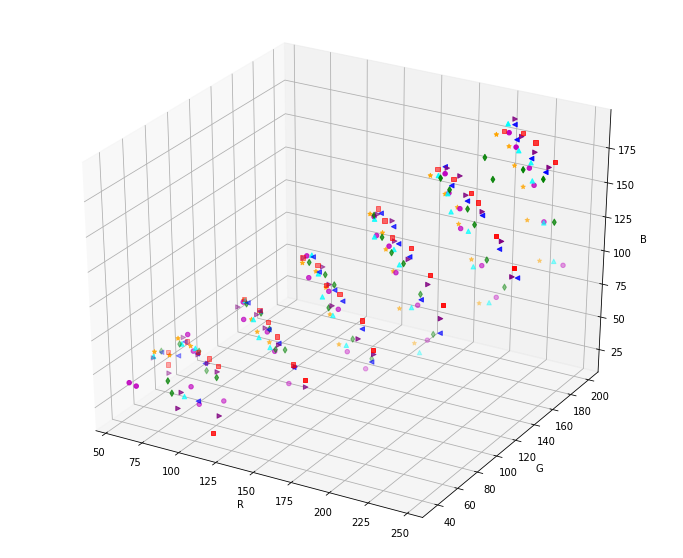}
    \caption{RGB to Munsell 3D Graph}
    \label{RGBtoMunsell}
    \end{subfigure}
       \hfill
    \begin{subfigure}[b]{0.52\textwidth}
    \centering
     \includegraphics[width=\textwidth]{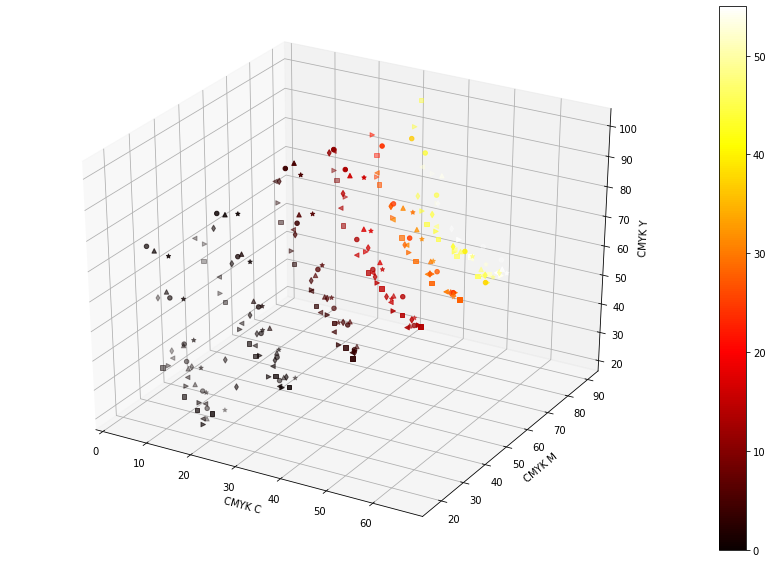}
    \caption{CMYK to Munsell}
    \label{cmyktomunsell}
    \end{subfigure}
       \hfill
    \begin{subfigure}[b]{0.32\textwidth}
     \centering
    \includegraphics[width=\textwidth]{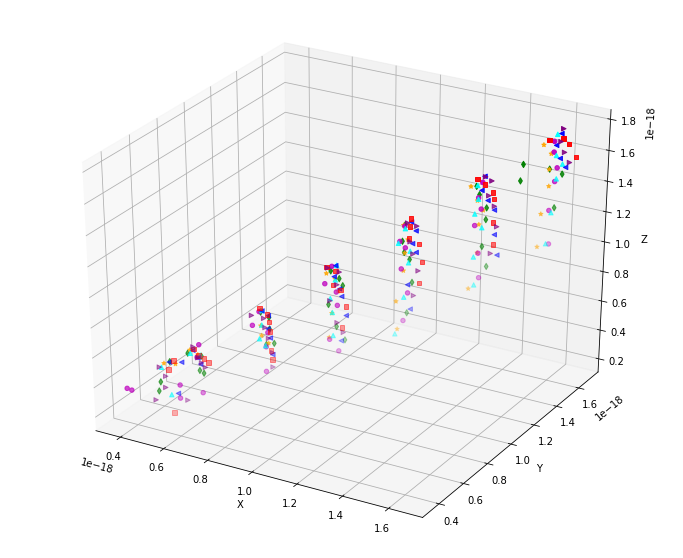}
    \caption{XYZ to Munsell}
    \label{xyztomunsell}
    \end{subfigure}
           \hfill
    \begin{subfigure}[b]{0.32\textwidth}
       \centering
    \includegraphics[width=\textwidth]{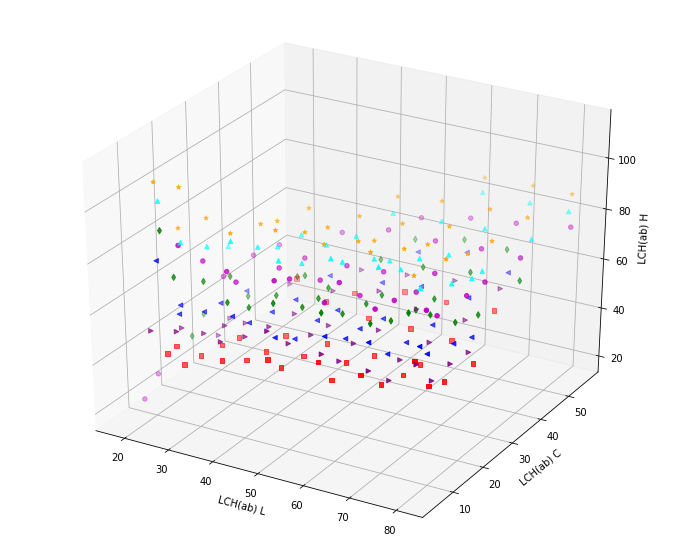}
    \caption{CIELCh to Munsell}
    \label{lchtomunsell}
    \end{subfigure}
          \hfill
    \begin{subfigure}[b]{0.32\textwidth}
        \centering
    \includegraphics[width=\textwidth]{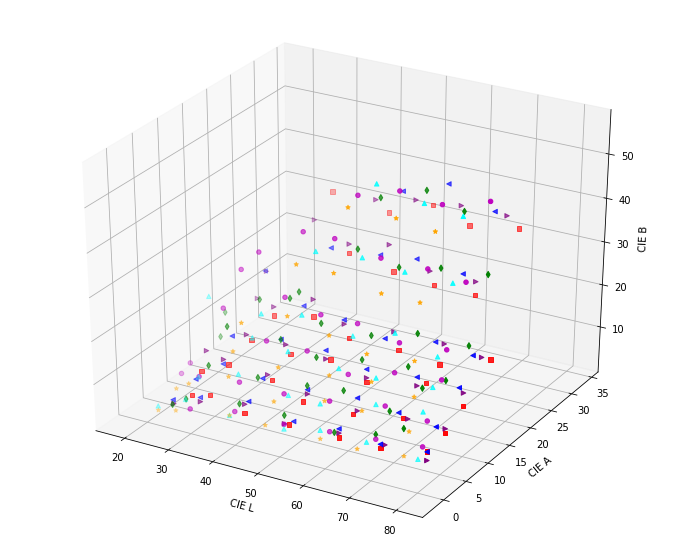}
    \caption{CIELab to Munsell}
    \label{labtomunsell}
    \end{subfigure}
    \caption{3D representation of Munsell to colorspace conversions}
    \label{graph}
\end{figure*}
A 3D representation of the Munsell Soil Color Chart to the chosen color space conversions is displayed in Fig. \ref{graph}. We are looking for a spread of values in these graphs to avoid `clumping'. This would avoid close matches in the color space that cause variations in readings due to sligh changes in values.
Fig. \ref{RGBtoMunsell} supports the use of a different color space as a medium as the `clumping' of values is evident in the higher and lower illumination values of RGB. This is particularly evident in the R-value of the RGB color space. The main reason for this is that several of the hues in the MSCC are reddish in color (10R, 2.5YR, 5YR). 
The CMYK system suffers from the same issues as the RGB system. As shown in Fig. \ref{cmyktomunsell}, the higher the illumination values the more `clumping' that occurs. The lower the K-value the higher the illumination of the colors. As this illumination increases to a K-value of \textless40, the MSCC values begin to become too close together to make this system unfeasible as a medium. 


XYZ color matching functions are simply transformations of the original RGB functions, using exactly the same data. The XYZ and RGB systems are completely interchangeable, and XYZ is often preferred for the lack of any negative components. As shown in Fig. \ref{xyztomunsell}, there is a significant `clumping' of results from the interpolations of XYZ to Munsell values. The extreme closeness of these values precludes them from use in the interpolation process from RGB to Munsell.


Fig. \ref{lchtomunsell} shows an even spread of Munsell values throughout the 3D graph. For this reason, there is good potential for the CIELCh color space to be utilised in the conversion of Munsell values. The separation according to the Hue values is also promising as they are clearly spread horizontally along the L (lightness) value. As illumination of samples is a cause for failing accuracy and repeatability the LCh color space could possibly be used in further applications to correct these luminance issues.


Pegalaja et al. \cite{Pegalajar2020} argues that the RGB measurements from a mobile phone obtained good results when converted to MSCC, but using a more stable color space, such as CIELAB, might improve the Munsell soil color approximations. Regardless, RGB values taken using an SPC need to be converted to CIELab before using the interpolation process~\cite{fan2017}. Conversely, an issue with CIELab is that it is not so easy to display directly as computer graphics and requires further conversion to RGB first.

CIELab, as discussed, may prove helpful in adapting values to compensate for illumination and other environmental variables once converted. Research concurs that CIELab values can be beneficial for correcting or altering color, especially when considering different levels of illumination \cite{fan2017}. In the two CIE color space models, the lightness variables L* showed good correlations with the MSCC conversion \cite{rossel2008}. CIELab is also the most widely used color space in the MSCC conversions, as it correlates closer to human vision than other color spaces and is uniform across the entire chromatic domain \cite{marq2018}. Most recent studies on soil color achieved excellent implementation using the CIELab system \cite{Stiglitz2017,Kirillova2018, Moritsuka2019, Kirillova2021, Soto2012}.


Fig. \ref{labtomunsell} displays an excellent spread of values across the 3D graph, similar to the CIELCh results. The separation of the Hue values is of particular use as the conversion of Munsell codes to RGB is not feasible at higher level values and chromas \cite{Aitkenhead2020}. The range of RGB values is not large enough to cover all of the possible Munsell soil color values. The same issue is not a problem for the CIELab color system, which can be used to derive the values of any Munsell soil color. As stated above, the RGB color space is unable to envelop the full spectrum of Munsell colors, which CIELab does. Munsell colors can be better represented when a larger number of parameters representing the color in different scales are used \cite{rossel2006}.
Our intention here was to determine which color scheme provided the best predictive results, informing future decisions when converting from RGB to the Munsell color system. As CIELab gives equal or better results than CIELCh, it is more appropriate to use the CIELab system than the others stated above \cite{Heil2022}.

\section{Color difference calculation techniques for MSCC}

There is a range of color difference calculations that can be utilised to calculate the conversion of one color space to another. In this section, an analysis was undertaken of the color difference calculation techniques used to convert CIELab values to the Munsell soil color values. As discussed previously, CIELab offers the best color space to act as a medium between the RGB values captured from an SPC to the MSCC. The color difference calculation techniques analysed in this section are CIEDE1976, CIEDE1994, CIEDE2000, and CMC. 

\begin{figure}[t!]
    \centering
    \includegraphics[width=0.45\textwidth]{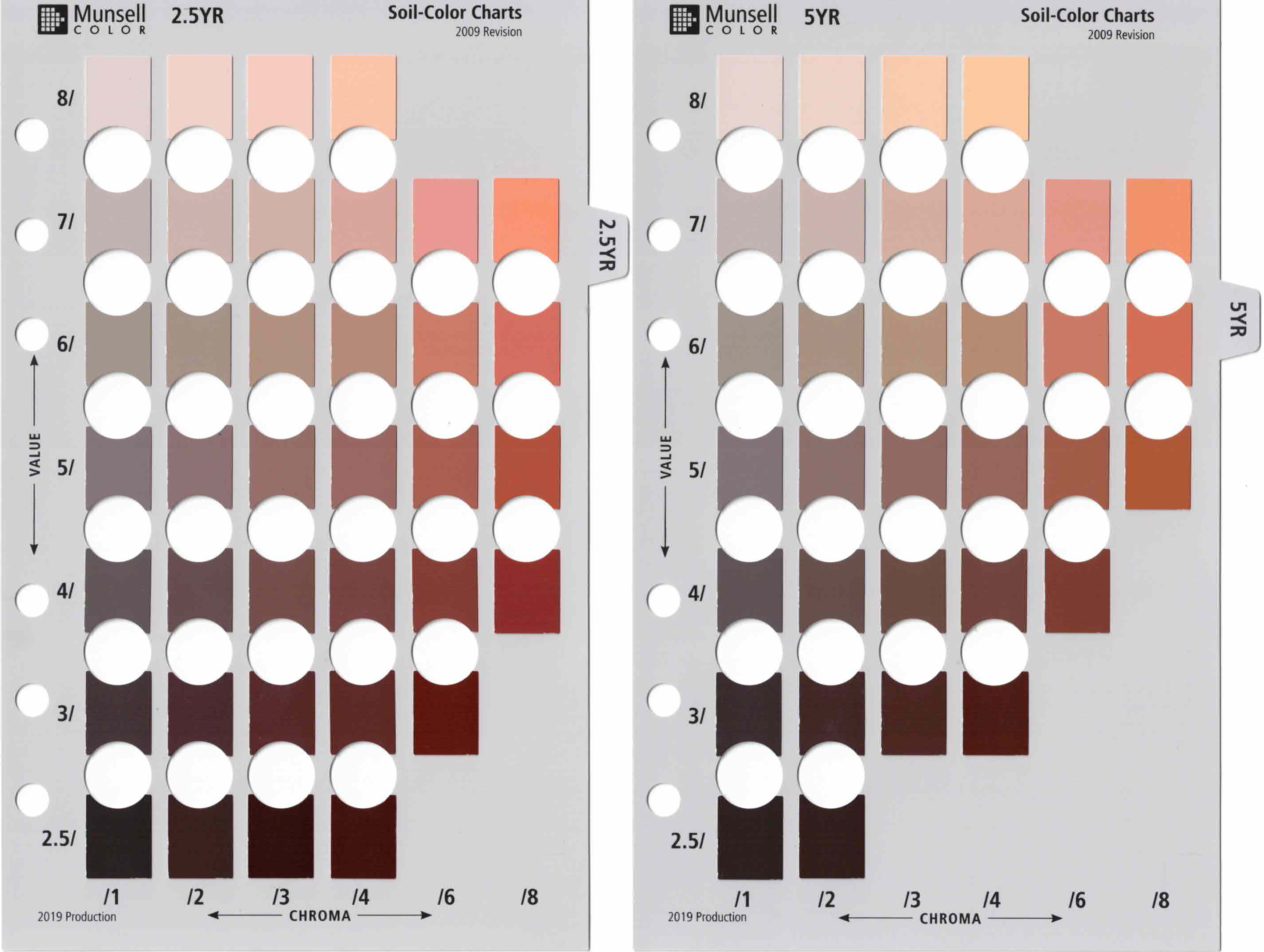}
    \caption{Munsell soil color chart 2.5YR and 5YR}
    \label{munsell2.5yr5yr}
\end{figure}

The most considerable similarities in Hue occur from 2.5YR to 5YR, especially in the lower chromas, as displayed in Fig. \ref{munsell2.5yr5yr}. In some cases, the interpolation results can vary significantly between hues due to the closeness of these values. For this reason, the analysis of the most appropriate color difference calculation will initially be analysed using the 2.5YR and 5YR MSCC hues. 

The difference is calculated based on Nix color sensor scan data in the CIELab color space and the Munsell values associated with this data. The CIELab values where captured from the MSCC using the Nix Pro 2 Color Sensor. Every color chip in the MSCC was analysed in triplicate, rotating the sensor after each analysis, totalling 437 unique Munsell color codes, 1311 analyses in total.

\begin{figure*}[!ht]
    \centering
    \includegraphics[width=0.89\textwidth]{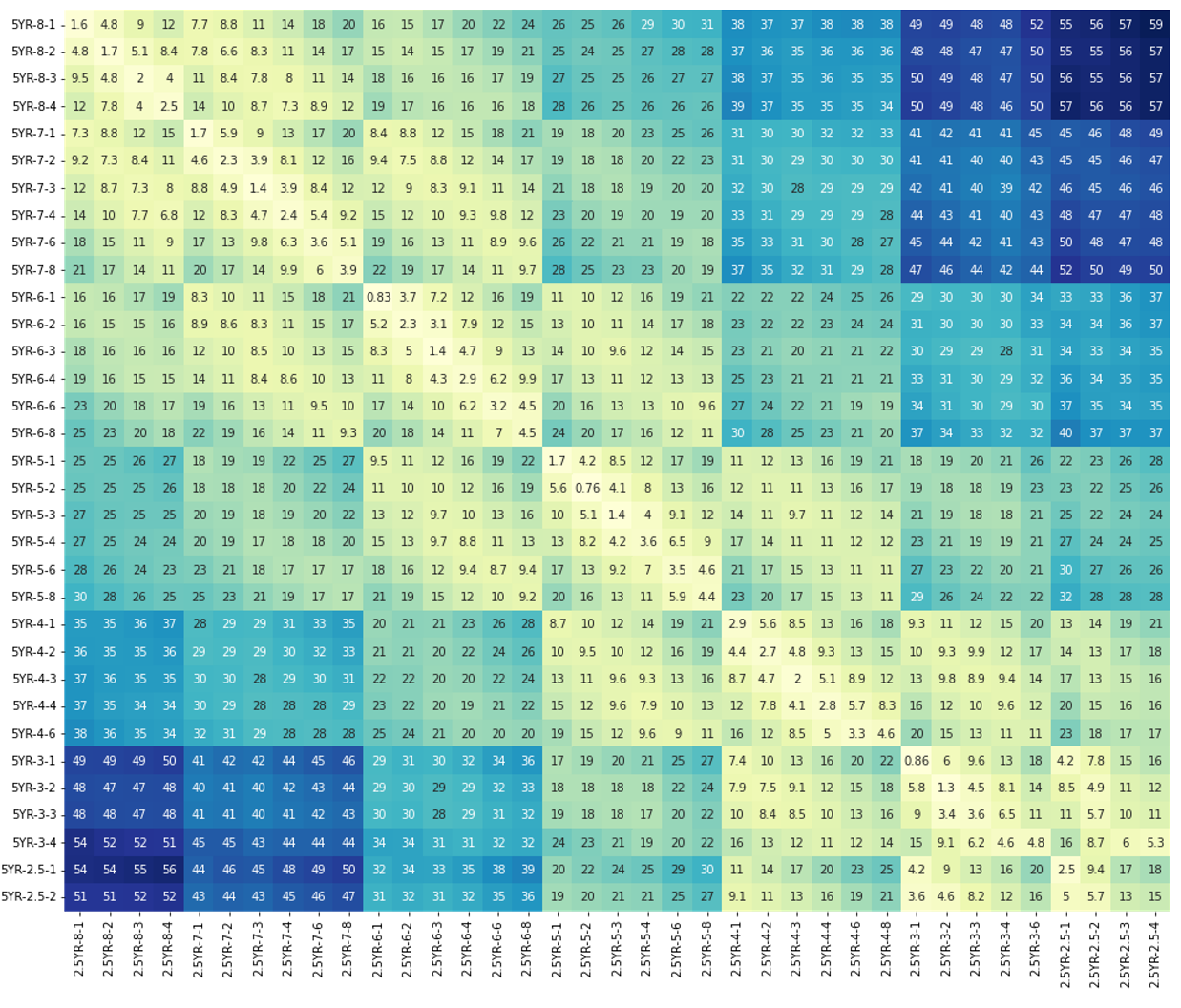}
    \caption{CIE2000 heatmap for CIELab to Munsell color chip similarities 2.5YR and 5YR}
    \label{2000heatmap}
\end{figure*}

Heatmaps have been created that display the closeness between the CIELab values and the MSCC values for the two hues from the MSCC in Fig. \ref{2000heatmap}. The closer the values the hotter the reading on the heatmaps. A heatmap for each color difference calculation technique is provided and discussed below. The lighter colors in these heatmaps indicate a closeness between the MSCC values. What we are looking for in these heatmaps is an even distribution of colors emanating from the centerline. The centerline is where the chip colors are the closest matches across hues, for example 2.5YR-8-1 and 5YR-8-1.

The D65 standard illuminant was chosen for use in all of the CIEDE calculations methods below as it corresponds roughly to the average midday light comprising both direct sunlight and indirect sunlight. For this reason, D65 is also known as the ``daylight illuminant". It is the fair assumption of this research that field samples for soil would be captured during daylight hours, supporting the use of D65.

CIEDE1976 is the standard CIE color difference method. This calculation measures the euclidean distance between two colors calculated in a three-dimensional space. The 1976 formula is the first formula that related a measured color difference to a known set of CIELAB coordinates. Euclidean distance is used to calculate the closeness between two color space values, enabling the researchers to calculate the Munsell distance between color values. In mathematics, Euclidean distance is the straight-line distance between two points in three-dimensional space.

The centerline values for the CIE1976 heatmap are a close match ranging from 7.6 to 1 with an average centerline value of 3.31 and a stdev 1.85. This is an excellent result due to the distribution of values becoming further apart emanating from the centerline. This leaves CIE1976 as a possible formula for calculating the closeness of CIELab to the MSCC.

CIEDE1994 is a more recent modification recommended and developed as the color difference formula for graphic arts and textile applications. ΔE (1994) is annotated as the LCh color space with tristimulus values of lightness, chroma and hue calculated from Lab coordinates. 

The 1994 formula performs poorly when considering the heatmap results. The centerline values are an extremely close match ranging from  0.48 to 6.1 with an average centerline value of 2.37 and a stdev 1.46. This is a discouraging result due to the distribution of values being uniformly similar across the entire table. CIEDE1994 is not a suitable candidate for calculating the closeness of CIELab to the MSCC. In defence of the 1994 color difference formula, it is meant to be used for the conversion of CIELab to CIELCh.

CIEDE2000 is a more recent modification of the CIE color difference formula. In this implementation KL = KC = KH = 1.0. Since the 1994 definition did not adequately resolve the perceptual uniformity issue, the CIE refined their definition, adding five corrections \cite{lindbloom2017}: (i) A hue rotation term (RT) to deal with the problematic blue region (hue angles in the neighbourhood of 275°), (ii) Compensation for neutral colors (the primed values in the LCh differences), (iii) Compensation for lightness (SL), (iv) Compensation for chroma (SC), and (v) Compensation for hue (SH).



As shown in Fig. \ref{2000heatmap}, the centerline values range from 5.7 to 0.76 with an average centerline value of 2.66 and a stdev of 1.21. This is a very good result due to the distribution of values becoming further apart emanating from the centerline. This asserts CIEDE1976 as a possible formula for calculating the closeness of CIELab to the MSCC.

For some of the DeltaE methods (CIEDE1994 and CIEDE2000), the color differences are not symmetric. This means that the difference between Color A and Color B may not be the same as that between Color B and Color A. In such cases, one color must be understood to be the reference or standard against which a sample color is compared.

The CMC method uses lightness and chroma weight. Developed by the Color Measurement Committee it is based on the CIELCh color space. The quasi metric has two parameters: lightness (l) and chroma (c), allowing the users to weigh the difference based on the ratio of l:c that is deemed appropriate for the application. As with CIE1994, this formula defines a quasimetric because it violates symmetry.

The CMC formula performs poorly when considering the heatmap results. The centerline values are an extremely close match ranging from  0.48 to 6.1 with an average centerline value of 3.94 and a stdev of 2.58. This is a discouraging result due to the distribution of values being too close and uniformly similar across the entire table. CMC is not a suitable candidate for calculating the closeness of CIELab to the MSCC.

\subsection{Color difference calculation using smartphones}\label{}

As the identified technology for use in this research is a smartphone camera (SPC) it is also necessary to test the capabilities of SPCs to identify the best color difference calculation method. The best method would be the calculation that can match the SPC images to the MSCC images captured with a Nix sensor with the highest accuracy. 

Table \ref{evalhue} and \ref{evalhvc} report the accuracy of matching results of 238 individual chip images of the MSCC (corresponding to '10YR', '7.5YR', '2.5Y', '5YR', '2.5YR', '5Y', '10R' pages) with the Nix sensor color value (in CIELab color domain). The values in the table display the percentage of chips that were correctly identified from the matching results. All of these images were taken in indirect sunlight at different times of the day using the same SPC for consistency. Set 1 was captured early morning, set 2 mid morning, set 3 midday, set 4 mid afternoon and set 5 late afternoon. The images of the MSCC were captured with the SPC for each set and the Nix sensor for comparison. 

Table \ref{evalhue} displays the percentage of matches that were correctly identified for the Hue only. From the results in this table, we can conclude that the CIEDE2000 based color difference calculation provides the highest accuracy in matching SPC images with the Nix sensor value for determining the Hue or similarly the correct page from the MSCC achieving 48.31\% accuracy.

\begin{table}[ht]
    \centering
    \caption{Accuracy of estimating Hue of mobile captured MSCC images (by comparing their color differences with Nix sensor) using different color difference calculation methods}
    \label{evalhue}
    \begin{tabular}{l*{9}{c}}
    \hline 
    Calculation & Set 1 & Set 2 & Set 3 & Set 4	 & Set 5\\ 
    \hline\hline
    CIE1976	& 29.83	& 7.56 & 46.63 & 42.85 & 47.05\\
    CIE1994	& 24.78	& 32.77	& 41.17	& 38.65	& 35.29\\
    CIE2000	& 31.09	& 39.49	& 46.21	& 47.89	& 48.31\\
    \hline
    \end{tabular}
\end{table}

Table \ref{evalhvc} displays the percentage of matches that were correctly identified for the Hue-Value-Chroma of each sample from the mobile phone to the MSCC chips captured by the Nix Color Sensor. Again the CIE2000 color space performs best with an overall accuracy of 15.96\%. This result is unacceptable for use in the solution when considering the potential inaccuracy it would generate. However, it is a good indication for once again confirming the selection of the CIEDE2000 calculation method.

\begin{table}[ht]
    \centering
    \caption{Accuracy of estimating Hue-Chroma-Value of mobile captured MSCC images (by comparing their color differences with Nix sensor) using different color difference calculation methods}
    \label{evalhvc}
    \begin{tabular}{l*{9}{c}}
    \hline 
    Calculation & Set 1 & Set 2 & Set 3 & Set 4	 & Set 5\\ 
    \hline\hline
    CIE1976	& 1.68 & 7.56 & 22.68	& 13.44	& 14.70\\
    CIE1994	& 1.26 & 0.84 & 2.10 & 1.68 & 2.94\\
    CIE2000 & 2.10 & 7.98 & 25.21 & 14.28 & 15.96\\
    \hline
    \end{tabular}
\end{table}

Overall CIEDE2000 is a better calculation method due to the excellent results from the CIELab color calculations of MSCC images compared to SPC images. Although the CIEDE1976 rates slightly better than CIE2000 in the Munsell color chip similarities (2.5YR and 5YR) heatmaps. These results are very close for both and the technology pinpointed for use is an SPC. The CIEDE2000 calculation will be the method that is adopted moving forward.

\section{Effect of smartphone cameras on estimating MSCC}

The feature of utmost importance required from an application of this type should be its precision and the repeatability of results. The color values returned from the soil samples must be as close to a match as possible to the MSCC. The need for an automated system of comparing soil samples to the MSCC becomes apparent when manually comparing the MSCC to soil samples. This manual comparison is very subjective, and results can widely vary between users \cite{Milotta2018b, Bloch2021}. Several other factors affect the accuracy of readings that have been identified when researching the possibility of utilising smartphones as an alternative to the MSCC human comparisons. The main two identified issues that affected accuracy included lighting, moisture variables and SPCs.
\begin{figure*}[t!]
\begin{subfigure}[b]{0.32\textwidth}
    \centering
    \includegraphics[width=\textwidth]{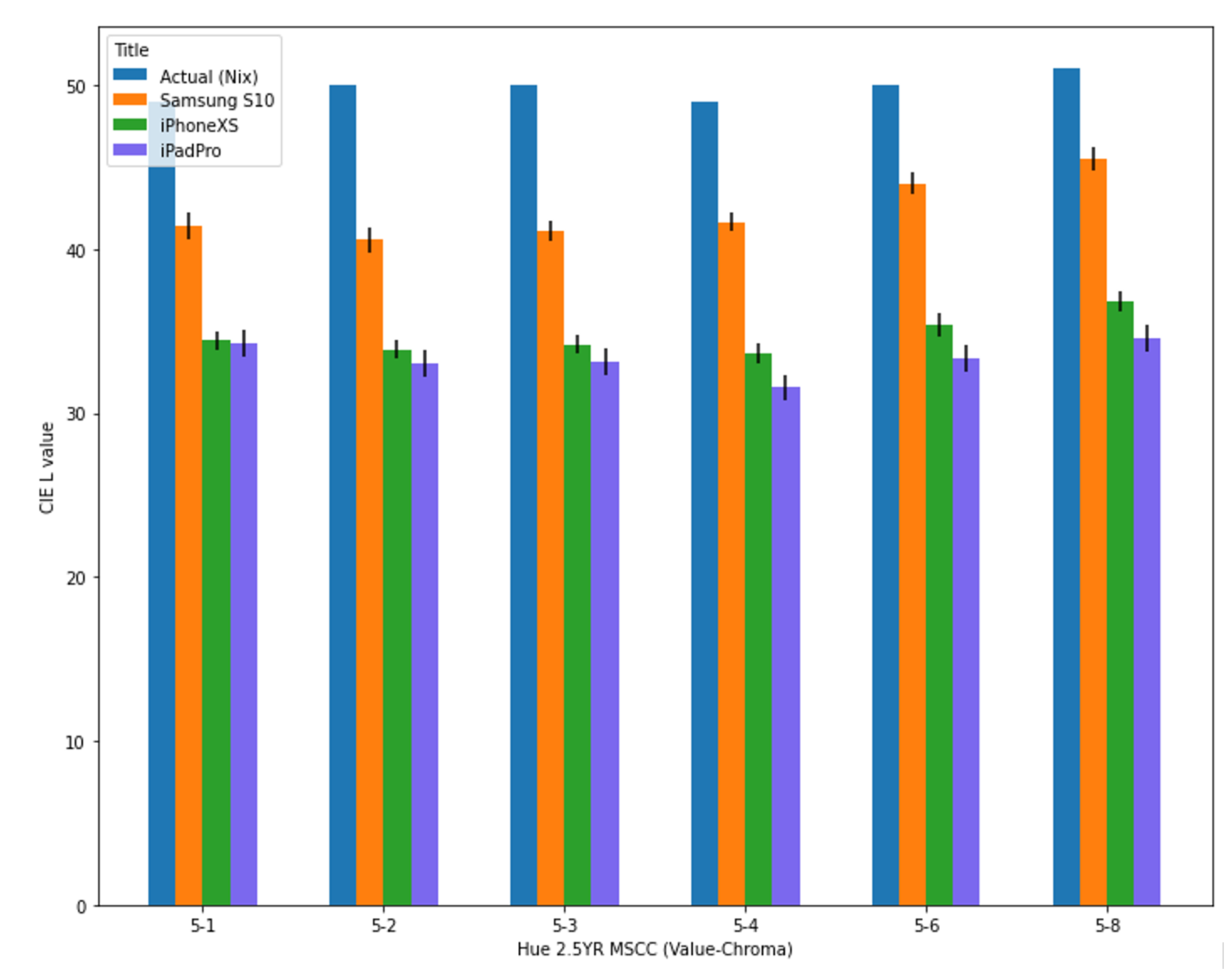}
    \caption{CIE L* value smartphone comparison}
    \label{L-spc}
\end{subfigure}
\hfill
\begin{subfigure}[b]{0.32\textwidth}
    \centering
    \includegraphics[width=\textwidth]{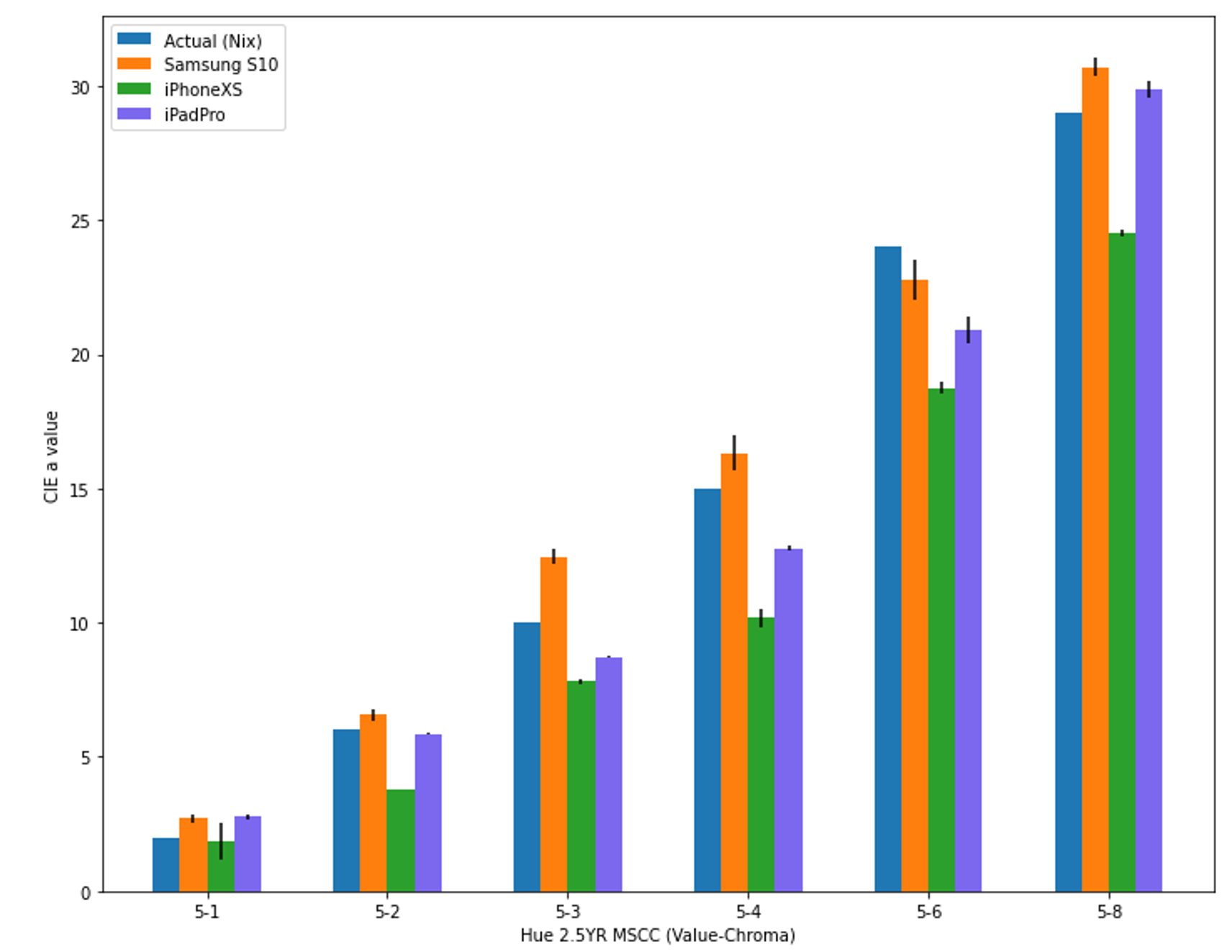}
    \caption{CIE a* value smartphone comparison}
    \label{a_spc}
\end{subfigure}
\hfill
\begin{subfigure}[b]{0.32\textwidth}
    \centering
    \includegraphics[width=\textwidth]{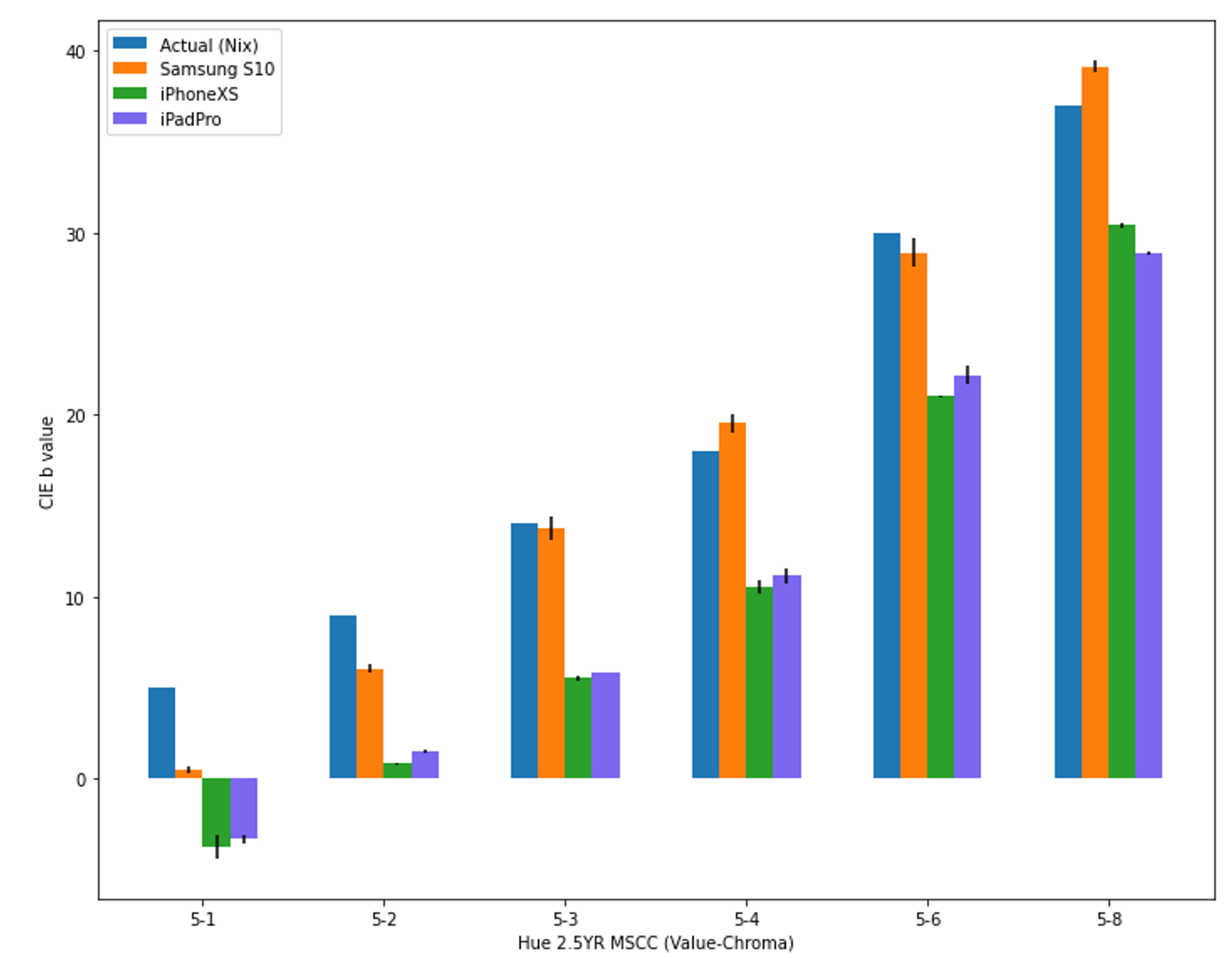}
    \caption{CIE b* value smartphone comparison}
    \label{b_spc}
\end{subfigure}
\caption{Samsung Galaxy S10, iPhone XS and iPad Pro CIELab value comparisons}
\label{labgraph}
\end{figure*}

When utilising SPCs, a variation in results from the same sample could be seen due to lighting and illumination issues, including the time of day, cloud cover, shadows from canopied areas, cracks, vegetation, soil texture, specular reflection and surface irregularities \cite{Bloch2021, fan2017, Fu2020, Gorthi2021, Mancini2020, Milotta2018a, Pegalajar2020, schmidt2021, Stiglitz2017, Turk2020}. Moisture levels of samples were also identified as factors that affect readings from SPCs \cite{Fu2020, Han2016, Stiglitz2017}. \cite{Han2016} explicitly tested the effects of moisture on samples. He concluded that increasing soil water content would ``cause the soil color deepening the differences in smartphone hardware affecting the accuracy of analysis results." However, the research from \cite{Stiglitz2017} established an opposite judgement finding that the ``Nix Pro Color Sensor determined the true color of a soil sample regardless of moisture content based on significant positive correlations between Nix Pro Color Sensor scans for samples in dry and moist conditions."

Irregular results can only partly be attributed to environmental difficulties of the type mentioned and whilst color space and color calculation methods are obvious factors to consider when determining the accuracy of predictions. There is, in addition, another more fundamental uncertainty with which the users of SPCs to calculate the values of the MSCC must consider. This appears to be based less on the method itself and more on the ability of the SPC to interpret colors under normal conditions. Let us amplify this somewhat more closely.

Different models of SPCs have different lenses and technologies to capture images. This presents researchers and users with the problem to ensure that results are independent of the devices used and the accuracy of readings is maintained regardless of the model of devices used. To illustrate the effects of different models and to gauge whether there are variations for different phones we have taken pictures of the MSCC pages using different phones at the same time of day. Six MSCC chips from Hue 5YR were captured (5-1, 5-2, 5-3, 5-4, 5-6, 5-8). Fig. \ref{labgraph}  images were captured using the Nix Pro 2 as the ``Actual” value along with a Samsung Galaxy S10, an iPhone XS and an iPad Pro. The CIELab values for each tristimulus (Lab) were then developed to analyse how SPCs have an impact on the perception of the color space.

The L* values from each device are displayed in Fig. \ref{L-spc}. Interpreting this table, we notice that the two Apple devices (iPhone XS and iPad Pro) are very similar. This is hypothetically because they both utilise the same camera modules (7 MP, f/2.2, 32mm). The Samsung S10 has a uniformly higher L* value with an average of 7.4. While the Nix Pro 2 has a much higher L* value than all the devices ranging from 5 to 16 values higher. This extreme difference can be attributed to the internal lighting from the built-in LED that the Nix Pro uses to illuminate samples and alleviate the issues associated with varying environmental lighting conditions. The ability of a smartphone to read environmental lighting and use this value to offset the differences between devices could be potentially implemented and warrants further investigation. 

In Fig. \ref{a_spc} the a* axis (green–red opponent colors) of the CIELab color space was analysed. There is a close alignment between the Samsung S10 and the actual values. Interestingly, the two apple devices are unaligned with the iPad Pro displaying higher values than the iPhone XS. The iPad Pro readings are closely aligned with the Samsung and the actual, whereas the iPhone XS displays an inability to accurately read the green-red values. This is a major disadvantage for the iPhone XS as many soils in the MSCC have a reddish hue. 


In Fig. \ref{b_spc} the b* axis (blue–yellow opponents) of the CIELab color space were analysed. Again the Samsung S10 provided the closest matching values to the actual. While the iPad Pro and iPhone XS were much lower than the actual readings, they showed a close relation to one another. This low reading from the Apple devices displays a difficulty in reading the actual blue-yellow opponents accurately. These blue-yellow opponents are essential in predicting the MSCC as several of the hues are yellowish in color. Again, a form of calibration would be beneficial to help align these values with the actual readings from the MSCC.


\section{Conclusion and future work}

The primary objective of this research was to provide a holistic analysis of the elements pertinent to creating a reliable, effective and repeatable automatic color prediction system using the Munsell soil color chart (MSCC). We designed our process to be specifically used in the field of soil science research however it may be modified to suit many professional applications such as food science, dentistry, printing, painting, and graphic design.  

The investigations above aim to establish the best methods for accurately and reliably predicting the Munsell soil color values. The most appropriate color space is CIELab due to its ability to cover all of the possible Munsell soil color values and the separation of the HVC to CIELab tristimulus when converted to a 3D graph. The color difference calculation that returned the best results was the CIEDE2000 method combined with the D65 illuminant. This issue with using an SPC is that several variables can affect the reliability of results, including different SPCs. For this reason, it was determined that a color correction technique is necessary to standardise colors across devices and offset potential environmental variables. 

Further research and experimentation are needed to improve the reliability of these results around the uncertainty of measuring soil color related to soil heterogeneity, this includes. Testing these methods on different smartphones to evaluate the effectiveness of the reference card. Utilising and evaluating the app using real-world soil samples in field conditions to judge the effects of variable lighting conditions, moisture levels and refraction/reflection angles. Additional comparisons against other methods advocated by researchers are needed to more precisely compare the results of this solution using the full range of MSCC chips.

Overall, this research has proven the best methods in developing a holistic solution for automating Munsell soil color predictions. 

\bibliographystyle{IEEEtran}
\bibliography{references}

\end{document}